\documentclass{article}

\usepackage{arxiv}

\usepackage{algorithm}
\usepackage{algorithmic}
\usepackage[utf8]{inputenc} 
\usepackage[T1]{fontenc}    
\usepackage{hyperref}       
\usepackage{url}            
\usepackage{booktabs}       
\usepackage{amsfonts}       
\usepackage{nicefrac}       
\usepackage{microtype}      
\usepackage{lipsum}		
\usepackage{graphicx}
\usepackage{makecell}
\usepackage{float}
\usepackage{doi}
\usepackage{xcolor}
\usepackage{makecell}
\usepackage[style=apa, sorting=nyt, backend=biber]{biblatex}
\title{A Comprehensive Study of LLM-Based Argument Classification: from LLAMA through GPT-4o to DeepSeek-R1}

\author{{\hspace{1mm}Marcin Pietroń} \\
	Faculty of Computer Science, \\ Electronics, and Telecommunications \\ AGH University of Krakow \\
	\texttt{pietron@agh.edu.pl} \\
     \And
 {\hspace{1mm}Filip Gampel} \\
     Faculty of Humanities
	 \\ AGH University of Krakow\\
	\texttt{fgampel@agh.edu.pl} \\
 \And
     {\hspace{1mm}Jakub Gomułka} \\
	Faculty of Humanities \\ AGH University of Krakow  \\
	\texttt{jgomulka@agh.edu.pl} \\
     \And
     {\hspace{1mm}Andrzej Tomski} \\
     Institute of Mathematics
	 \\ University of Silesia\\
	\texttt{andrzej.tomski@us.edu.pl} \\
    	\And
       {\hspace{1mm}Rafał Olszowski} \\
	Faculty of Humanities \\ AGH University of Krakow  \\
	\texttt{rolszowski@agh.edu.pl} \\
}

\renewcommand{\shorttitle}{\textit{arXiv} Template}

\begin{document}
\maketitle

\begin{abstract}

Argument mining (AM) is an interdisciplinary research field that integrates insights from logic, philosophy, linguistics, rhetoric, law, psychology, and computer science. It involves the automatic identification and extraction of argumentative components, such as premises and claims, and the detection of relationships between them, such as support, attack, or neutrality. Recently, the field has advanced significantly, especially with the advent of large language models (LLMs), which have enhanced the efficiency of analyzing and extracting argument semantics compared to traditional methods and other deep learning models. There are many benchmarks for testing and verifying the quality of LLM, but there is still a lack of research and results on the operation of these models in publicly available argument classification databases.
This paper presents a study of a selection of LLM's, using diverse datasets such as Args.me and UKP. The models tested include versions of GPT, Llama, and DeepSeek, along with reasoning-enhanced variants incorporating the Chain-of-Thoughts algorithm. The results indicate that ChatGPT-4o outperforms the others in the argument classification benchmarks. In case of models incorporated with reasoning capabilities, the Deepseek-R1 shows its superiority. However, despite their superiority, GPT-4o and Deepseek-R1 still make errors. The most common errors are discussed for all models. In addition, the differences between models are shown.
Finally, a model with rephrased prompts and certainty is proposed. It improves the argument recognition accuracy, but it is still not perfect. 
To our knowledge, the presented work is the first broader analysis of the mentioned datasets using LLM prompt algorithms. The work also shows some weaknesses of known prompt algorithms in argument analysis, while indicating directions for their improvement.
The added value of the work is the in-depth analysis of the available argument datasets and the demonstration of their shortcomings.


\end{abstract}

\keywords{Argument mining \and Transformers \and  Large Language Models \and Cognitive intelligence \and NLP}

\section{Introduction}



Argument mining (AM) is a research field that spans multiple disciplines, including logic and philosophy, linguistics, rhetoric, law, psychology, and computer science. Although the study of argumentation theory and the use of logical reasoning to support claims and conclusions has a long history, the application of data science techniques to automate these processes is a relatively new advancement. The ability to automatically extract arguments and their relationships from various input sources is crucial across many domains. Several early approaches to what is now known as argument mining began to emerge around 2010. During this time, initial methods for extracting various connotations of arguments from natural language documents were proposed. For example, \textcite{teufel2009towards} introduced the concept of argumentative zoning in scientific articles, while \textcite{mochales2011argumentation} developed a method to detect arguments within legal texts. In the past decade, AM has emerged as a central focus within the field of artificial intelligence (\cite{cabrio2018five}; \cite{schaefer2022gercct}; \cite{lawrencereed}), due to its ability to conjugate representational needs with user-related cognitive models and computational models for automated reasoning \parencite{lippi2016argumentation}. As a subfield of Natural Language Processing (NLP) and computational linguistics, AM focuses on automatically identifying, extracting, and analyzing argumentative structures within natural language texts, including recognizing core components of arguments, such as claims and evidence \parencite{park-cardie-2018-corpus}. 


Early research in AM focused mainly on edited texts (\cite{moens}, \cite{levy-etal-2014-context}, \cite{stab-gurevych-2014-annotating}).
However, more recent studies have expanded the focus to include user-generated content from a variety of platforms, such as debate portals and social media sites such as Facebook or Twitter \parencite{dusmanu-etal-2017-argument}. This shift highlights the increasing relevance and applicability of argument mining in analyzing and understanding various types of discourse in different online environments. Initially, argument structures were often represented using trees or tree-like models.
Before the advent of BERT and other Transformer-based models, Support Vector Machines (SVMs) and neural networks were pivotal in AM, leveraging their pattern recognition and classification capabilities to identify and analyze argumentative structures in the text. 
The transformer-based architecture and its pre-trained models have marked a significant breakthrough in the field of NLP. Conversely, the advent of new Large Language Models (LLMs) has opened up additional opportunities. These LLM capabilities can be useful in analyzing arguments more efficiently.




It should be mentioned that language models like Llama or GPT are increasingly being deployed for general problem solving across a wide range of tasks but are still confined to token-level, left-to-right decision-making processes during inference. This means they can fall short in tasks that require exploration.
Therefore, exploration strategies were designed as Chain of Thoughts (\cite{wei2022chain}), Thread of Thoughts (\cite{zhou2023thread}) or Tree of Thoughts (\cite{yao2023tree}). These algorithms enable exploration over coherent units of text that serve as intermediate steps toward problem solving. For this reason, a new generation of models with a built-in reasoning  process has recently been created (e.g. Deepseek-R1, \cite{guo2025}).
Some of the most popular benchmarks for testing LLM models and prompt algorithms concentrate on knowlegde choice tests (most genuine grade-school level, multiple-choice science questions e.g. ARC easy and ARC challenge, Boolq), logic games (e.g. Game of 24, Mini Crosswords) and code or text generation (e.g. Creative Writing). There is a lack of tests on argument classification datasets. 
In the case of arguments, syntax, semantic, and logical rules are difficult to define, not counting the context of argument construction. It makes the process of argument classification different problem from well-known benchmarks. 




In AM, the task is typically divided into three main subtasks based on their argumentative complexity. The first sub-task is the identification of argument components, which involves distinguishing argumentative propositions from non-argumentative ones. This step enables the segmentation of the input text into arguments, allowing subsequent subtasks to be performed. The second subtask is the identification of clausal properties, where the focus is on identifying premises or conclusions within the argumentative propositions. The third and final subtask is the identification of relational properties, which involves analyzing two different argumentative propositions at the same time to determine the type of relationship between them. This task often involves deciding whether a given text serves as an argument for, an argument against, or is unrelated to the input thesis. A classifier developed in this way can aid in constructing a model to extract arguments from the text (Fig.\ref{fig:architecture}).

This paper aims to explore the capabilities of various LLMs in addressing the challenge of argument classification. To achieve this, we used two different datasets from well-known AM projects: UKP and Args.me. Each of these datasets includes argument classifications conducted by human annotators, who assigned labels based on the type of argument. 
The key tasks in argument mining require high-quality annotated corpora to train and evaluate the performance of automated approaches. 
This work presents a qualitative analysis of the available arguments datasets.
In our study, we attempted to automatically classify arguments from these datasets, and then compared the results with the annotations previously made by humans. 

We addressed the following research questions:

\textbf{RQ1. How does a particular syntax of prompts affect the quality of argument classification?}


\textbf{RQ2. What are the performance differences between different language models and how does the size of the LLMs increase the capabilities of argument classification?}

\textbf{RQ3. How do prompting and reasoning algorithms improve argument classification by LLMs?}

\textbf{RQ4. How can the certainty-based multiprompt LLM model improve the precision of classification arguments?}

\textbf{RQ5. What types of error do LLMs make in automatic argument classification?}

\textbf{RQ6. What are the shortcomings of existing annotated datasets commonly used to train or evaluate networks and models in argument classification and how they should be improved?}


To our knowledge, this study presents one of the first evaluations of modern LLMs on argument classification benchmarks. In addition, it serves as an introduction to the development of an efficient prompt algorithm for the analysis of arguments in natural language. The work also shows some weaknesses of some well-known prompt algorithms in argument analysis, while indicating directions for their improvement.
The added value of the work is the in-depth analysis of the available argument datasets and the demonstration of their shortcomings.

\section{Related works}
\label{sec:headings}

Argument classification (AC) is a specialized subtask in the broader field of AM. It concentrates on categorizing the identified elements of an argument into predetermined classes. These classes often include differentiating claims and premises, or identifying whether a component supports or opposes the argument. Additionally, argument classification can entail assessing the nature of the argument, such as determining its type or evaluating its quality (e.g., whether it is a strong or weak argument). This task is integral to constructing a structured representation of arguments. By clarifying the role and relationship of each component, it enhances our understanding of the dynamics within an argumentative discourse \parencite{lippi2016argumentation}. Argument classification (AC), as a specialized subtask within AM, was described by \textcite{daxenberger2020argumentext}, \textcite{lippi2016argumentation}, \textcite{dusmanu-etal-2017-argument} and many others. According to these authors, AC involves categorizing the identified components of an argument into predefined categories, such as differentiating between claims and premises or determining the stance of an argumentative component, whether supporting or opposing. AC may also include assessing the type of argument or evaluating its quality (e.g., strong versus weak arguments). This task is essential for constructing a structured representation of arguments, as it aids in understanding the role of each component within the argument and how different components interact with one another.

Various data science techniques, that utilize natural language processing, have proven effective in AC and, more broadly, in AM. Initially, argument structures were often represented using trees or tree-like models, facilitating computation due to the availability of tree-based parsing techniques. However, real-world arguments frequently deviate from these idealized structures. More recently, researchers have shifted towards exploring non-tree-based argument structures in argument mining. Before the advent of BERT and other Transformer-based models, Support Vector Machines (SVMs) and neural networks were pivotal in AM, leveraging their pattern recognition and classification capabilities to identify and analyze argumentative structures in the text. 
Architectures such as Recurrent Neural Networks (RNN), e.g. Long Short-Term Memory (LSTM), and Convolutional Neural Networks (CNN) have been used to incorporate contextual information into machine decision-making processes. \textcite{niculae2017argument}  introduced the first non-tree model for argument mining, using a factor graph model combined with structured Support Vector Machines and Bidirectional Long Short-Term Memory. SVMs have been widely used in argument mining due to their effectiveness in binary classification tasks, which are well-suited to identifying whether a particular sentence or phrase is an argumentative component (e.g., claim vs. non-claim). Subsequently, \textcite{galassi2018argumentative} utilized LSTMs and residual network links to predict the connections among argument components. LSTM based approach for argument classification is also presented in \cite{ref_stab18}. These LSTM-based solutions have limits and can achieve up to 45\% accuracy in the UKP dataset, \cite{ref_stab18, li2020empirical}.


While SVMs and neural networks significantly contributed to the development of AM, they had limitations, such as the need for extensive feature engineering and difficulties in capturing long-distance dependencies in text. The introduction of BERT (Bidirectional Encoder Representations from Transformers) and subsequent Transformer-based models revolutionized AM, \cite{pietron2024efficient, li2020empirical}. The Transformer architecture, featuring its self-attention mechanism, was initially introduced by \textcite{vaswani2017} as a solution to the increasing computational and memory demands of recurrent neural networks (RNNs), which were considered state of the art at the time. 
Research on how transformers work, especially the BERT model \parencite{devlin-etal-2019-bert}, has attracted significant research interest. Fine-tuning large pre-trained Transformer-based models resulted in remarkable performance gains across a wide range of tasks. Several recent studies have employed transformer-based models for argument mining. \textcite{reimers2019classification} leveraged contextual word embeddings such as BERT and ELMo to significantly enhance argument/non-argument classification and proposed methods for argument clustering. \textcite{chakrabarty2019ampersand} introduced a BERT-based model for argument component classification and relation detection within persuasive online conversations. Going further, \textcite{chen2021bert} is using pre-trained BERT based models for predicting arguments where the structure forms a directed acyclic graph. Moreover, \textcite{ruiz2020transformer} present an analysis of the behavior of transformer-based models (i.e., BERT, XLNET, RoBERTa, DistilBERT and ALBERT) when predicting argument relations, and evaluate the models in five different domain specific corpora, with the objective of finding the less domain dependent model. The work presented in \cite{li2020empirical} shows that DistilBERT achieves 54.2\% and the BERT precision is around 57.7\% in the UKP benchmark. In \cite{pietron2024efficient} authors present the BERT incorporated with ChatGPT-4. The proposed architecture significantly outperforms other ML-based solutions. It reports 89.5\% accuracy in the Args.me and 68.5\% on the UKP benchmark. Recently, new Transformer-based architectures were designed, e.g. BLOOM (\cite{lescao2023}), Llama (\cite{touvron2023}) and its new versions, GPT-4o and DeepSeek-R1 (\cite{guo2025}). There is a lack of works which study the performance of these models in argument mining, especially those with reasoning capabilities. Chain of thought is an approach that simulates human-like reasoning processes by delineating complex tasks into a sequence of logical steps towards a final solution. This methodology reflects a fundamental aspect of human intelligence, offering a structured mechanism for problem-solving. This technique can be applied in prompt or can be incorporated in LLM reasoning-based model.
LLM benchmarks mostly concentrate on e.g. Game of 24, Creative Writing, or Mini Crosswords. 
\cite{liu2024}, \cite{lan2024}
\cite{zhang2024}, \cite{xia2024}, \cite{zhou2025}

Given the challenges posed by fully automated AM methods, an interesting research direction has emerged in exploring hybrid approaches that combine the efforts of human annotators with AI. 
The automated AM methods often struggle to determine whether two arguments express the same viewpoint (\cite{chakrabarty2019ampersand}; \cite{daxenberger2017essence}), and reliance on a limited set of labeled data can lead to the exclusion of minority opinions, thereby creating a bias towards more popular or frequently repeated arguments. For example, a hybrid method called HyEnA \parencite{vandemeer2024hybrid} employs a sampling algorithm that guides human annotators individually through an opinion corpus. Then, an intelligent merging strategy helps annotators combine their results into clusters of arguments, integrating both manual and automatic labeling. The presented work tries to analyze argument datasets and LLM errors which can help to improve training process (e.g. improving the training data) or setup rules for the manual annotations.   

\section{Datasets}

In our research, we decided to conduct a comparative study using two datasets containing different argument corpora, each of which was developed by a different research team. These were: the UKP corpus \parencite{stab2018cross}, and the Args.me corpus \parencite{ajjour2019data}. These datasets have gained recognition in recent argument mining research, but they have not yet been studied comparatively or with the latest large language models such as Llama, DeepSeek or GPT-4o. The previous studies, including \cite{bar-haim-etal-2017-stance}, \cite{boltuzic2014back},  \cite{stab2018cross} were limited to a research on a single dataset only, and covered pre-transformer-based NLP technologies.

Both datasets used in our study share fundamental structural characteristics of argumentation, namely:

\begin{itemize}
    \item Presence of a thesis – a central conclusion or topic around which the argumentation is constructed; 
    \item  Provision of a list of premises – each labeled as either supporting or opposing the thesis. 
\end{itemize}

The form and quality of argument descriptions vary between datasets; for instance, some include off-topic entries, non-arguments, or rephrased conclusions in place of genuine premises.

The first dataset used in our study was the UKP dataset \parencite{stab2018cross}, which comprises a corpus of arguments derived from online comments on eight controversial topics: abortion, cloning, the death penalty, gun control, minimum wage, nuclear energy, school uniforms, and marijuana legalization. This corpus includes over 25,000 instances. The statistics for the UKP datasets are presented in Tables \ref{statistc_of_ukp} and \ref{statistc_of_ukp_}. The sentences were independently annotated by seven individuals recruited through the Amazon Mechanical Turk (AMT) crowdsourcing platform. For each classification, an agreement level was required, with Cohen’s kappa ($\kappa$) set at 0.723, surpassing the commonly accepted threshold of 0.7 for reliable results \parencite{carletta1996assessing}. The classification labels used were: (1) supporting argument (FOR), (2) opposing argument (AGAINST), and (3) non-argument (NO ARGUMENT).

\begin{table}[H]
\centering
\begin{tabular}{|c|c|c|c|c|c|}
\hline
\textbf{} & \textbf{abortion} & \textbf{cloning}  & \textbf{death penalty} &  \textbf{marijuana legalization} \\ \hline
\textbf{NON}     &      2282     &   1472                            &   1918 & 1160 \\ \hline
\textbf{PRO}     &       634         &     702                            &   424  & 535\\ \hline
\textbf{CON}     &       766        &     825                            &  1036  & 574\\    \hline

\end{tabular}
\caption{UKP database statistics part 1}
\label{statistc_of_ukp}
\end{table}

\begin{table}[H]
\centering
\begin{tabular}{|c|c|c|c|c|c|}
\hline
\textbf{} & \textbf{gun control} & \textbf{minimum wage}  & \textbf{nuclear energy} & \textbf{school uniforms} \\ \hline
\textbf{NON}     &      1846     &   1323                             &   2051 & 1734 \\ \hline
\textbf{PRO}     &       775          &     564                             &   591  & 545\\ \hline
\textbf{CON}     &       650        &     541                            &  831  & 729\\    \hline

\end{tabular}
\caption{UKP database statistics part 2}
\label{statistc_of_ukp_}
\end{table}

The second dataset used in our study was the Args.me corpus (version 1.0, cleaned), provided by \textcite{ajjour2019data}. The distribution of this corpus is presented in Table \ref{statistc_of_argsme}. This dataset consists of arguments collected from four debate portals in mid-2019: Debatewise, IDebate.org, Debatepedia, and Debate.org. The arguments were extracted using heuristics specifically designed for each debate portal. The sub-datasets used in our simulations were Idebate.org, Debatepedia, and Debatewise, which together contain 48,798 arguments. Debatepedia has the highest number of arguments, while IDebate.org shows the least disproportion between PRO and CON arguments, with the greatest disproportion observed in Debatepedia. Args.me is the newest and most extensive dataset among all those utilized in our study. The annotations in this dataset include conclusions and premises, which are further categorized into PRO premises (arguments supporting the conclusion) and CON premises (arguments opposing the conclusion).

All calculations were performed on statistically representative subsamples of 200 entries per dataset partition (eight subsamples of 200 for UKP and three subsamples of 200 for Args.me), owing to financial constraints.







\begin{table}[H]
\centering
\begin{tabular}{|c|c|c|c|c|}
\hline
\textbf{} & \textbf{idebate.org} & \textbf{debatepedia}  & \textbf{debatewise} \\ \hline
\textbf{conclusions}     &      5011        &       10314                         &  5992 \\ \hline
\textbf{premises}     &     13248            &      21197                            &  14353 \\ \hline
\textbf{PRO premises}     &     6701           &      15791                            &   8514 \\ \hline
\textbf{CON premises}     &     6547            &     5406                            &   5839 \\ \hline

\end{tabular}
\caption{Args.me database statistics}
\label{statistc_of_argsme}
\end{table}




\section{LLM based argument classification}

\subsection{System architecture}

\begin{figure}[h]
    \centering
    \includegraphics[width=15cm]{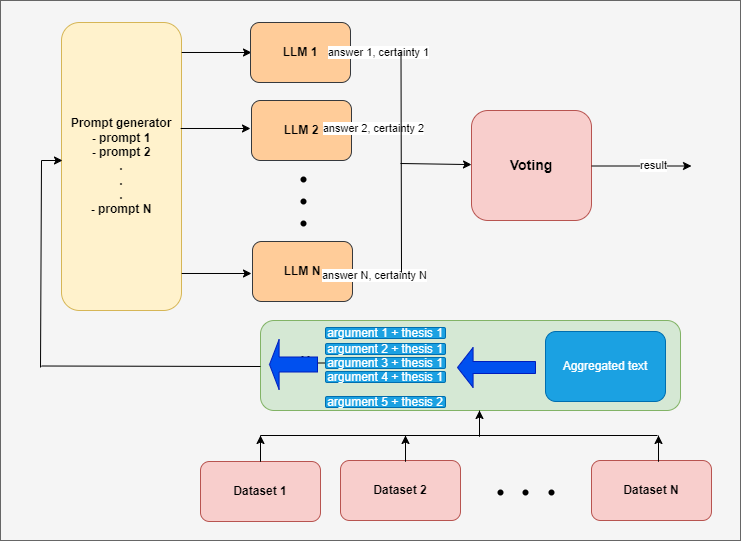}
    \caption{System architecture.}
    \label{fig:architecture}
\end{figure}

The general architecture of the approach presented is described in Fig. \ref{fig:architecture}. The input to the system is a bunch of the multiple datasets. All datasets are aggregated, and then chunks of data (parts of text with possible arguments) are sent to the Prompt Generator module (additionally, the thesis can be included in a chunk). The Prompt Generator creates the set of prompts related to the domain to which the input chunks belong. If the thesis is empty (e.g. in UKP dataset) then the Prompt Generator creates the thesis based on the argument domain. The number of prompts can be predefined or extended to any amount. After that bunch of LLM models (e.g. Llama 1B, Llama 13B, Llama 70B, GPT-4) process the generated prompts.
Each LLM gives an answer and certainty as an output. The LLM outputs are inputs to the voting module. The Voting module incorporates a few voting strategies to generate the final output of the system. Voting strategies in the final stage can minimize the errors generated by single models. The whole system set up the ensemble strategy, which is based on a set of pre-trained LLMs.

\subsection{Large Language Models architecture}

The main module of the models is the Transformer, which was proposed in \parencite{vaswani2017} as an architecture to efficiently model sequential data. These models consist of stacked blocks incorporated with self-attention layers:



\begin{equation}
T_{\Theta}(X) = {t_{\theta_L}(t_{\theta_{L-1}}...(t_{\theta_{0}}(X)))}
\label{eq:model}
\end{equation}

where the $\Theta$:
\begin{equation}
\Theta = \{\theta_{0}, \theta_{1},...,\theta_{L}\}
\label{eq:theta_}
\end{equation}

is a list of parameters $\theta_{i}$ for each Transformer block in a sequence $t_i$. Each tensor $\theta_{i}$ is a set of linear layers' weights in a single transformer block. 
Each transformer block is a combination of the attention layer (each attention has three linear layers) and the feed forward module (two linear layers). At the end of each block the normalization layer and linear layer is performed.

Each attention layer receives the embeddings. The $i$-th token is mapped via linear transformations to a key $k_i$, query $q_i$ and value $v_i$. The $i$-th output of the self-attention layer is given by weighting the values $v_j$ by the normalized dot product between the query $q_i$ and other keys $k$:


\begin{equation}
    z_{i}=\sum_{j=1}^{n}softmax(\{<q_{i}, k_{j'}>\}_{j'=1}^{n})_j \cdot v_{j}
\end{equation}

The $T_{\Theta}$ gives an output with the same number of tokens as the input $X$ (generates embedding for each token).

In our research, the OpenAI API is utilized to engage ChatGPT-4, ChatGPT-4o and Deepseek-R1. The open source model used in the presented research is Llama in the following versions: Llama 3.1 8B and 70B, Llama 3.2 1B and 3B and Llama 3.3 70B. We also study a distilled version of Deepseek R1 based on Llama 3.3 70B. There are some slight differences between Transformer blocks between Llama and GPT (e.g. grouped multi-query attention vs. multi-head attention).  
Deepseek-R1 is the only fully reasoning model in this set of models.



\subsection{Argument Mining prompts}

In the first stage, the models were tasked with analyzing and categorizing arguments without any prior specific training on a similar task or dataset related to argument mining. This approach allowed us to assess the models' abilities to classify arguments without prior training on specific argumentative datasets. 

The series of structured queries aimed at determining the stance of various arguments relative to specified theses were formulated. The queries were designed to evaluate not only the binary classification capability of the models (whether an argument supports or opposes a thesis), but also to explore their ability to quantify certainty. 
The specific questions (prompts) posed to the models included:
\begin{enumerate}
    \item Assess if the argument supports or opposes the presented thesis, with responses limited to 'for, against, or no argument' (Args.me dataset).
    \item Asking for the stance ('for' for support, 'against' for opposition, 'no argument' if non-argumentative) of a statement with respect to a debate topic characterized by a single keyword or a thesis (UKP dataset).
    \item Request an assessment of the certainty of the classification, expressed as a percentage (both UKP and Args.me dataset).

\end{enumerate}

Due to the fact that many prompt algorithms improve the performance of LLM models on many benchmarks, an attempt was made to examine some of them. The first is a modification of Rephrase and Respond (RaR). RaR allows LLMs to rephrase and expand questions posed by humans and provide responses in a single prompt. This approach serves as a simple yet effective motivation method to improve performance \parencite{deng2023rephrase}. For this purpose, four representative examples were developed for each set. The examples are variations of the original prompt and are used to query the models:

\begin{enumerate}
    \item \textbf{Prompt P1}: \textit{"Is the sentence: "..." an argument for or against "...", or is it no argument? Return one of the expressions: "For", "Against" or "No argument", without any additional commentary."} 
    
    This featured a query structure, with the thesis formulated in the most straightforward manner possible (e.g., "For or against the death penalty," "For or against the minimum wage"). The model was required to present the response as the stance of the argument in relation to the thesis, using verbal labels: \textit{"For"}, \textit{"Against"}, or (in the UKP database only) \textit{"No argument"}.
    
    \item \textbf{Prompt P2}: \textit{The thesis is: "..." Indicate if the argument "..." is for this thesis (F), against this thesis (A), or neutral (N). Please respond with only one letter: F, A, or N, without any additional commentary.} 
    
    This included a more elaborate formulation of the thesis, serving as a reference point for the premises in the arguments (e.g., \textit{"The death penalty should be allowed,"} \textit{"The minimum wage is justified and should be increased"}). Additionally, the response format was restricted to the \textbf{F/A/N} coding scheme.
    
    \item \textbf{Prompt P3}:  \textit{"In the context of the ongoing public debate, evaluate whether the text "..."  represents an argument supporting or opposing "...", or whether it does not qualify as an argument at all. Respond with one of the expressions: "For", "Against" or "No argument".}
    
    This prompt required consideration of the most complex relationships, instructing the model to classify the argument while taking into account the \textbf{context of the ongoing public debate}, with the response format containing the stance of the argument expressed in verbal form, as in \textbf{P1}.
    
    \item \textbf{Prompt P4}: \textit{Is the sentence: "..." an argument for (F) or against (A) "..." or is it no argument (N)? Return a single letter: F, A, or N, without any additional commentary.}
    
    This represented the simplest variant, where the thesis was formulated in the same straightforward manner as in \textbf{P1}, but the response format was restricted to the \textbf{F/A/N} coding scheme, as in \textbf{P2}.

\end{enumerate}

\begin{figure}[h]
    \centering
    \includegraphics[width=8cm]{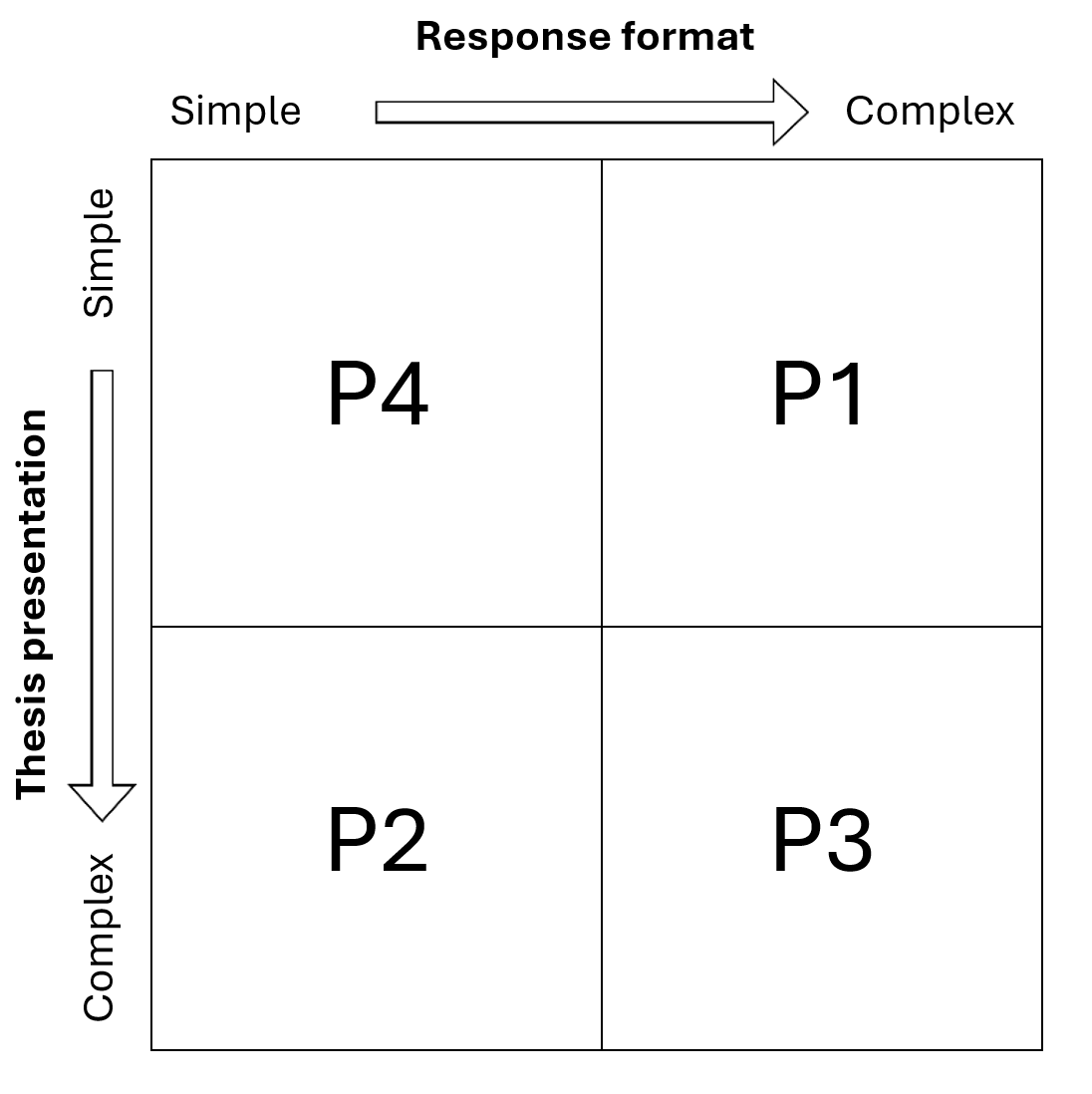}
    \caption{Overview of prompt complexity used in the study.}
    \label{fig:prompts}
\end{figure}

The comparison of prompt complexity is presented in Fig. \ref{fig:prompts}, which shows the complexity of the thesis presentation (vertical axis) and the response format (horizontal axis). The prompts shown here were used with the UKP database, while minimally adapted versions (without the "No argument" option) were used for Args.me.

The next prompting technique is Chain of Thoughts (CoT) technique mirrors human reasoning and problem-solving through a coherent series of logical deductions. Chain of thought prompting is an approach in artificial intelligence that simulates human-like reasoning processes by delineating complex tasks into a sequence of logical steps towards a final resolution. This methodology reflects a fundamental aspect of human intelligence, offering a structured mechanism for problem solving \parencite{wei2022chain} - prompting the LLM to first analyze and summarize with additional information/context step by step, before answering
\parencite{zhou2023thread}.
For this purpose, a specially built prompt was developed to indicate the model's inference path. The CoT prompt is defined as follows:

\textbf{Prompt CoT}: \textit{Is the sentence: "..." an argument for or against "...", or is it no argument?
Solve the argument classification problem. Think through the problem step by step to solve it. At each step, you have to figure out:
\begin{itemize}
    \item the step number, 
    \item the sub - question to be answered in that step,
    \item the thought process of solving that step, and
    \item the result of solving that step.
\end{itemize}
Respond in the following markdown table format for each step: "| step | subquestion | process | result |". The result should be one of the expressions: "For", "Against" or "No Argument", without any additional commentary.}

The third approach used is a few-shot prompt, which can be used as a technique to enable learning in the context, where we provide demonstrations in the prompt to steer the model to better performance \parencite{ma2023fairness}. In our case, the prompts were formed by choosing as an example one sentence for each category (For, Against, No Argument) for each of the eight UKP datasets.




Finally, an additional prompt was used so that the model could assess the certainty of its answer. For this purpose, an algorithm was built that is based on several prompts and the certainty of the answer, which is defined as:

\begin{equation}
T_{\Theta}(P)=\sum c_{i} \cdot T_{\Theta}(p_{i}) 
\label{eq:model}
\end{equation}

\begin{algorithm}
\begin{algorithmic}[1]
\REQUIRE{$L$ -- set of LLM models}
\REQUIRE{$\gamma$ -- threshold}
\REQUIRE{$\Omega$ -- list of prompts}
\REQUIRE{$\Omega_c$ -- list of prompts}
\STATE{$A \gets \emptyset$}
\FOR{$l$ $\textbf{in}$ $L$}
\STATE{$\phi \gets \emptyset$}
\STATE{$\kappa \gets \emptyset$}
\STATE{$\delta_f \gets$ 0}
\STATE{$\delta_a \gets$ 0}
\STATE{$\delta_n \gets$ 0}
\FOR{$\omega$ $\textbf{in}$ $\Omega$}
\STATE{$\phi \gets \phi \cup l(\omega)$}
\IF{$l(\omega) == \textbf{'For'}$}
\STATE{$\delta_f = \delta_f + 1$}
\ELSIF{$l(\omega) == \textbf{'Against'}$}
\STATE{$\delta_a = \delta_a + 1$}
\ELSE
\STATE{$\delta_n = \delta_n + 1$}
\ENDIF
\STATE{$\kappa \gets \kappa \cup l(\omega_c)$}
\ENDFOR
\STATE{$\Delta = max(\delta_f, \delta_a, \delta_n)$}
\IF{$(\delta_f == \delta_a == \Delta) \vee (\delta_f == \delta_n == \Delta) \vee (\delta_n == \delta_a == \Delta)$}
\STATE{$out_l \gets 0$}
\FOR{$\omega$ $\textbf{in}$ $\Omega$}
\STATE{$out_l = out_l + \kappa[i] \cdot l(\omega)$}
\ENDFOR
\ELSE
\STATE{$out=argmax(\delta_f, \delta_a, \delta_n)$}
\ENDIF
\STATE{$A \gets A \cup out_l$}
\ENDFOR
\RETURN{$A$}
\caption{Evaluation of the models with few prompts}
\label{alg:p_model}
\end{algorithmic}
\end{algorithm}

The algorithm is presented in Algorithm \ref{alg:p_model}.
The algorithm in the main loop processes subsequent LLMs (from line 2 to 29). The algorithm initializes data structures (from line 3 to 7). Next, the processed model is queried with subsequent prompts from the set (line 9). During processing, the types of individual responses are counted (lines from 10 to 16). For each type of prompt, an additional query is made to the model about the certainty of its previous answer (line 17). In the case of a predominance of the selected answer, the model accepts it as the final one. The line 20 contains a condition that, if met (the number of different most frequent answers is equal), the certainty of individual answers is taken into account. Based on the certainty, the final answer in the line 23 is determined (line 21 to 24). 

\section{Results and discussion}

In this section results from running a wide spectrum of large language models on different argument datasets are described. Additionally, the efficiency of argument classification of different prompt strategies is shown.  

\subsection{Few-shot prompting}

Table \ref{tab:few-shot} shows the results of a few-shot experiment on a selection of datasets used in the study. We used the four UKP datasets to prompt the five Llama models with a variant of prompt P1:

\textbf{Few-shot prompt}: \textit{Is the sentence: [sentence] an argument for or against [topic], or is it no argument? Return one of the expressions: "For", "Against" or "No argument", without any additional commentary. Here are some examples: \\
Sentence: [E1] Answer: For \\
Sentence: [E2] Answer: Against \\
Sentence: [E3] Answer: No argument} \\

Here E1, E2 and E3 are example statements as described in the previous section. They can be found in Appendix A.

\begin{table}[H]
\centering
\caption{Few-shot vs Zero-shot results}
\label{tab:few-shot}
\small
\begin{tabular}{lccccc}
\toprule
\textbf{Dataset} & \makecell{Llama 1B \\ 3-shot | 0-shot } & \makecell{Llama 3B \\ 3-shot | 0-shot } & \makecell{Llama 8B \\ 3-shot | 0-shot } & \makecell{LLama 3.1 70B \\ 3-shot | 0-shot } & \makecell{Llama 3.3 70B \\ 3-shot | 0-shot }\\
\midrule
UKP – Abortion              & 54.8 | \textbf{57.0} & 51.8 | \textbf{57.0} & \textbf{68.0} | 67.5 & \textbf{74.6} | 72.0 & \textbf{78.2} | 73.5 \\
UKP – Cloning               & \textbf{53.3} | 51.0 & \textbf{62.4} | 59.0 & 62.4 | \textbf{78.0} & \textbf{82.7} | 80.0 & 80.7 | \textbf{82.0} \\
UKP – Death Penalty         & 56.9 | \textbf{58.0} & 57.4 | \textbf{61.0} & 60.9 | \textbf{69.0} & 70.6 | \textbf{72.5} & 70.1 | \textbf{72.5} \\
UKP – Gun Access            & \textbf{54.8} | 49.5 & 50.3 | \textbf{60.5} & 49.2 | \textbf{66.0} & 70.1 | \textbf{73.0} & 69.0 | \textbf{71.0} \\
UKP – Marijuana Legalisation & \textbf{55.3} | 50.5 & 53.8 | \textbf{60.0} & 55.3 | \textbf{67.0} & \textbf{76.1} | 73.5 & \textbf{75.1} | 72.0 \\
UKP – Nuclear Energy       & \textbf{59.4} | 58.5 & \textbf{65.5} | 58.5 & 66.5 | \textbf{75.0} & 71.6 | \textbf{75.5} & 68.5 | \textbf{73.0} \\
UKP – School Uniforms       & \textbf{55.8} | 44.0 & 59.9 | \textbf{67.0} & 67.0 | \textbf{78.0} & 83.8 | \textbf{86.0} & \textbf{85.3} | 85.0 \\
UKP – Minimum Wage          & \textbf{50.3} | 46.0 & 59.4 | \textbf{68.5} & 65.5 | \textbf{75.5} & 71.1 | \textbf{80.5} & 64.0 | \textbf{80.0} \\
\midrule
\textbf{UKP - Total}        & \textbf{55.1} | 51.8 & 57.6 | \textbf{61.4} & 61.9 | \textbf{72.0} & 75.1 | \textbf{76.6} & 73.9 | \textbf{76.1} \\
\bottomrule
\end{tabular}
\end{table}

The results show that in most cases the few-shot approach does not improve the 0-shot results. Only for the smallest model the technique shows on average an improvement. Moreover, in some cases the difference in favor of the 0-shot approach is significant. This is likely due to the fact that, the examples presented are not representative enough for the wide range of arguments occurring in the dataset. The conclusions from these results are that the model, in order to be able to improve the results with a high probability, should have well-chosen examples for a specific group of arguments. Such a solution could be dynamically selecting examples using the RAG technique.

\subsection{RaR and CoT prompting}

The experiments carried out demonstrate that the models evaluated — starting from LLaMA 3.1 70B and above — show high accuracy in the classification tasks of arguments. In contrast, significantly lower accuracy is observed with smaller models, such as LLaMA 1B to 8B. Table~\ref{tab:model_results} presents the average accuracy scores for all the prompts used in the evaluation (from P1 to P4).

\begin{table}[H]
\centering
\caption{Average model accuracy in argument classification on the UKP and Args.me datasets}
\label{tab:model_results}
\small
\begin{tabular}{llllllllll}
\toprule
{} & Llama 1B & Llama 3B & Llama 8B & Llama   & Llama   & DS 70B & DS R1 & GPT4 & GPT4o \\
{} & {}       & {}       & {}       & 3.1 70B & 3.3 70B & {}     & {}    & {}   & {} \\
\midrule
UKP - abortion & 36.4 & 53.6 & 61.7 & 69.3 & 72.1 & 76.1 & 79.5 & 80.0 & \textbf{82.0} \\
UKP - cloning & 36.3 & 52.0 & 73.8 & 80.3 & 80.3 & 82.0 & 80.0 & 79.5 & \textbf{83.9} \\
UKP – death penalty & 36.5 & 51.0 & 64.3 & 70.5 & 72.8 & 77.0 & 78.9 & 77.4 & \textbf{81.2} \\
UKP – gun access & 36.0 & 53.8 & 60.4 & 66.4 & 66.9 & 74.4 & 76.3 & 75.1 & \textbf{78.1} \\
UKP – marijuana legalisation & 37.3 & 54.9 & 67.1 & 73.3 & 72.0 & 76.0 & 78.0 & 78.3 & \textbf{81.5} \\
UKP – nuclear energy & 39.6 & 52.9 & 69.5 & 73.4 & 72.4 & 77.6 & 80.2 & \textbf{81.2} & 78.9 \\
UKP – school uniforms & 34.1 & 59.6 & 72.8 & 83.9 & 83.4 & 83.8 & 85.1 & 81.5 & \textbf{85.7} \\
UKP – minimal wage & 36.4 & 57.5 & 71.2 & 79.0 & 79.8 & 78.1 & 79.8 & 81.1 & \textbf{82.6} \\
\textbf{Avg. UKP result} & 36.6 & 54.5 & 67.9 & 74.4 & 74.9 & 78.1 & 79.7 & 79.3 & \textbf{81.7} \\
\textbf{Avg. UKP (Certainty)} & 41.8 & 60.6 & 71.4 & 76.3 & 77.3 & 81.3 & 79.9 & 81.0 & \textbf{84.3} \\
\textbf{Avg. UKP (CoT)} & 27.6 & 56.4 & 64.7 & 76.6 & 78.8 & 78.1 & \textbf{80.1} & - & 78.5 \\
Args.me - idebate & 46.4 & 63.7 & 74.6 & 87.6 & 88.9 & 89.3 & \textbf{92.9} & 83.3 & 89.7 \\
Args.me - debatepedia & 33.0 & 59.4 & 69.1 & 84.9 & 85.1 & 91.0 & \textbf{94.4} & 82.8 & 86.2 \\
Args.me - debatewise & 38.7 & 57.4 & 73.9 & 84.3 & 85.4 & \textbf{86.3} & 83.1 & 80.3 & 85.8 \\
\textbf{Avg. Args.me result} & 39.4 & 60.1 & 72.5 & 85.6 & 86.5 & 88.9 & \textbf{90.1} & 82.1 & 87.2 \\
\textbf{Avg. Args.me (Certainty)} & 41.9 & 63.7 & 77.5 & 88.1 & 88.6 & 89.8 & 90.4 & 88.4 & \textbf{90.6} \\
\textbf{Avg. Args.me (CoT)} & 45.1 & 55.7 & 70.6 & 80.7 & 81.0 & 88.9 & \textbf{90.1} & - & 89.5 \\
\bottomrule
\end{tabular}
\end{table}

\begin{table}[H]
\centering
\caption{Average model precision in argument classification on the UKP and Args.me datasets}
\label{tab:model_results2}
\small
\begin{tabular}{llllllllll}
\toprule
{} & Llama 1B & Llama 3B & Llama 8B & Llama   & Llama   & DS 70B & DS R1 & GPT4 & GPT4o \\
{} & {}       & {}       & {}       & 3.1 70B & 3.3 70B & {}     & {}    & {}   & {} \\
\midrule
UKP - abortion & 34.6 & 50.9 & 55.2 & 68.9 & 71.0 & 73.7 & 71.2 & \textbf{80.3} & 78.7 \\
UKP - cloning & 27.5 & 56.6 & 74.8 & 81.0 & 80.3 & 81.1 & 79.2 & 84.2 & \textbf{84.8} \\
UKP – death penalty & 37.3 & 45.6 & 57.0 & 66.4 & 69.4 & 72.2 & 75.0 & \textbf{80.0} & 79.6 \\
UKP – gun access & 33.8 & 54.3 & 56.3 & 63.4 & 63.6 & 70.5 & \textbf{76.8} & 71.3 & 74.7 \\
UKP – marijuana legalisation & 33.2 & 54.9 & 66.3 & 72.7 & 72.0 & 74.5 & \textbf{83.7} & 82.1 & 81.9 \\
UKP – nuclear energy & 33.0 & 53.3 & 65.0 & 71.7 & 70.3 & 74.8 & 74.2 & \textbf{79.8} & 75.8 \\
UKP – school uniforms & 33.2 & 57.7 & 69.8 & 82.4 & 82.2 & 82.3 & 80.4 & 83.0 & \textbf{85.0} \\
UKP – minimal wage & 34.0 & 64.1 & 72.0 & 79.1 & 79.2 & 77.9 & 77.4 & \textbf{83.6} & 81.7 \\
\textbf{Avg. UKP result} & 33.3 & 54.7 & 64.6 & 73.2 & 73.5 & 75.9 & 77.2 & \textbf{80.5} & 80.3 \\
\textbf{Avg. UKP (Certainty)} & 35.5 & 59.3 & 68.2 & 74.7 & 75.9 & 79.7 & 77.2 & 82.4 & \textbf{82.6} \\
Args.me - idebate & 55.1 & 65.5 & 76.6 & 88.4 & 89.2 & 90.1 & 89.6 & 86.7 & \textbf{90.3} \\
Args.me - debatepedia & 37.8 & 60.0 & 68.7 & 81.3 & 81.6 & 88.3 & \textbf{98.5} & 79.9 & 82.8 \\
Args.me - debatewise & 45.5 & 57.5 & 75.1 & 83.6 & 84.6 & 85.7 & \textbf{91.6} & 82.7 & 83.6 \\
\textbf{Avg. Args.me result} & 46.1 & 61.0 & 73.5 & 84.5 & 85.2 & 88.0 & \textbf{93.2} & 83.1 & 85.5 \\
\textbf{Avg. Args.me (Certainty)} & 61.8 & 62.7 & 76.0 & 86.6 & 87.0 & 89.3 & \textbf{93.3} & 86.9 & 88.9 \\
\bottomrule
\end{tabular}
\end{table}

\begin{table}[H]
\centering
\caption{Average model recall in argument classification on the UKP and Args.me datasets}
\label{tab:model_results3}
\small
\begin{tabular}{llllllllll}
\toprule
{} & Llama 1B & Llama 3B & Llama 8B & Llama   & Llama   & DS 70B & DS R1 & GPT4 & GPT4o \\
{} & {}       & {}       & {}       & 3.1 70B & 3.3 70B & {}     & {}    & {}   & {} \\
\midrule
UKP - abortion & 35.4 & 50.8 & 54.6 & 73.9 & 75.3 & 76.1 & \textbf{88.6} & 74.0 & 81.1 \\
UKP - cloning & 34.0 & 51.9 & 76.1 & 84.9 & 85.5 & 83.6 & \textbf{87.5} & 77.5 & 83.6 \\
UKP – death penalty & 34.6 & 45.4 & 58.2 & 70.4 & 71.7 & 74.5 & 77.8 & 72.4 & \textbf{80.6} \\
UKP – gun access & 35.7 & 51.5 & 56.7 & 66.4 & 67.0 & 73.8 & \textbf{86.3} & 69.2 & 75.1 \\
UKP – marijuana legalisation & 34.5 & 53.2 & 67.2 & 79.6 & 78.8 & 79.1 & 76.6 & 77.6 & \textbf{81.7} \\
UKP – nuclear energy & 35.6 & 50.8 & 64.5 & 78.4 & 77.7 & 77.7 & \textbf{87.8} & 78.5 & 80.8 \\
UKP – school uniforms & 32.0 & 54.1 & 67.2 & 85.1 & 84.8 & 83.1 & \textbf{90.2} & 76.8 & 84.6 \\
UKP – minimal wage & 34.6 & 60.6 & 72.4 & 82.7 & 83.6 & 78.4 & \textbf{85.7} & 77.3 & 82.1 \\
\textbf{Avg. UKP result} & 34.5 & 52.3 & 64.6 & 77.7 & 78.0 & 78.3 & \textbf{85.1} & 75.4 & 81.2 \\
\textbf{Avg. UKP (Certainty)} & 35.3 & 61.2 & 69.2 & 79.6 & 80.1 & 81.2 & \textbf{85.2} & 77.2 & 84.2 \\
Args.me - idebate & 46.4 & 63.7 & 74.7 & 87.6 & 88.9 & 89.3 & \textbf{92.0} & 83.3 & 89.7 \\
Args.me - debatepedia & 45.1 & 60.7 & 70.1 & 82.9 & 85.9 & \textbf{90.5} & 90.1 & 82.7 & 84.9 \\
Args.me - debatewise & 46.0 & 57.1 & 73.6 & 84.4 & \textbf{85.9} & 85.4 & 82.9 & 77.6 & 84.7 \\
\textbf{Avg. Args.me result} & 45.8 & 60.5 & 72.8 & 85.0 & 86.5 & \textbf{88.4} & 88.3 & 81.2 & 86.5 \\
\textbf{Avg. Args.me (Certainty)} & 51.3 & 62.9 & 77.7 & 87.3 & 88.8 & 89.5 & 88.6 & 88.9 & \textbf{90.8} \\
\bottomrule
\end{tabular}
\end{table}

\begin{table}[H]
\centering
\caption{Average model F1 in argument classification on the UKP and Args.me datasets}
\label{tab:model_results4}
\small
\begin{tabular}{llllllllll}
\toprule
{} & Llama 1B & Llama 3B & Llama 8B & Llama   & Llama   & DS 70B & DS R1 & GPT4 & GPT4o \\
{} & {}       & {}       & {}       & 3.1 70B & 3.3 70B & {}     & {}    & {}   & {} \\
\midrule
UKP - abortion & 25.8 & 45.6 & 53.4 & 68.5 & 70.4 & 73.6 & \textbf{79.0} & 75.7 & 78.7 \\
UKP - cloning & 24.0 & 47.7 & 73.2 & 81.0 & 80.03 & 81.5 & 83.2 & 79.0 & \textbf{83.6} \\
UKP – death penalty & 24.4 & 37.7 & 52.6 & 64.5 & 66.6 & 72.4 & 76.4 & 73.7 & \textbf{79.0} \\
UKP – gun access & 26.3 & 47.4 & 54.2 & 62.9 & 62.8 & 71.2 & \textbf{81.3} & 69.7 & 74.3 \\
UKP – marijuana legalisation & 25.3 & 47.7 & 64.6 & 73.3 & 72.1 & 75.7 & 79.1 & 69.7 & \textbf{81.2} \\
UKP – nuclear energy & 26.2 & 44.4 & 63.0 & 72.8 & 71.4 & 75.7 & 80.4 & 78.4 & \textbf{81.2} \\
UKP – school uniforms & 23.2 & 48.8 & 65.9 & 83.0 & 82.5 & 82.5 & \textbf{85.1} & 78.7 & 84.5 \\
UKP – minimal wage & 26.1 & 56.2 & 70.7 & 79.3 & 79.6 & 77.6 & 81.3 & 79.1 & \textbf{81.4} \\
\textbf{Avg. UKP result} & 25.2 & 46.9 & 62.2 & 73.2 & 73.2 & 76.2 & \textbf{80.7} & 76.6 & 80.0 \\
\textbf{Avg. UKP (Certainty)} & 34.4 & 58.4 & 68.3 & 75.3 & 76.3 & 80.2 & 81.0 & 79.2 & \textbf{83.2} \\
Args.me - idebate & 40.6 & 62.6 & 73.8 & 87.6 & 88.9 & 89.3 & \textbf{89.8} & 83.8 & 89.7 \\
Args.me - debatepedia & 25.9 & 55.8 & 65.0 & 81.9 & 82.7 & 89.2 & \textbf{94.11} & 80.7 & 83.4 \\
Args.me - debatewise & 33.2 & 55.5 & 72.5 & 83.7 & 84.9 & 85.5 & \textbf{87.0} & 79.6 & 83.7 \\
\textbf{Avg. Args.me result} & 33.3 & 58.0 & 70.4 & 84.4 & 85. & 88.0 & \textbf{90.3} & 81.4 & 85.6 \\
\textbf{Avg. Args.me (Certainty)} & 31.7 & 61.7 & 76.0 & 86.8 & 87.6 & 89.3 & \textbf{90.6} & 87.5 & 89.6 \\
\bottomrule
\end{tabular}
\end{table}

As observed, there are noticeable differences in model performance across the various datasets. In the case of the UKP datasets, GPT-4o demonstrates the highest classification accuracy. GPT4o performs best on all subsets except the nuclear energy subset for which GPT4 performs best. GPT4o achieves the highest mean value of 84.3\%, which is significantly better than Deepseek-R1's mean value of 80.1\%.  

The situation is different with the Args.me data sets, where the best performing model is DeepSeek-R1 (90.1\%), surpassing both GPT-4 (82.1\%) and GPT-4o (87.2\%), as well as all versions of LLaMA. However, in the case of the version that takes into account the certainty values, the GPT4o model achieves the highest value (90.6\%) and slightly outperforms Deepseek-R1. In the case of precision values (Tab. \ref{tab:model_results2}), the results for the UKP subsets are mixed. The best in each subset are achieved by the GPT4, GPT4o and Deepseek-R1 models. Finally, the average value is achieved by the GPT4 and GPT4o models. In the case of the Args.me set, the precision values are definitely the highest for the Deepseek-R1 model. The highest recall values for UKP subsets are achieved by Deepseek-R1. Only for death penalty and marijuana legalization is GPT4o better. In the case of Args.me the highest average recall value is achieved by GPT4o (90.8\%). Finally, the highest F1 values are achieved by GPT4, GPT4o and Deepseek-R1 for individual UKP subsets. The average F1 value for the UKP set with certainty is the highest for GPT4o (83.2\%), while without certainty for Deepseek-R1 (80.7\%). The average value of Deepseek-R1 F1 (90. 6\%) for the Args.me dataset is the best and it is 1\% higher than the second best result achieved by GPT4o (89.6\%).

\begin{figure}[h]
    \centering
    \includegraphics[width=16 cm]{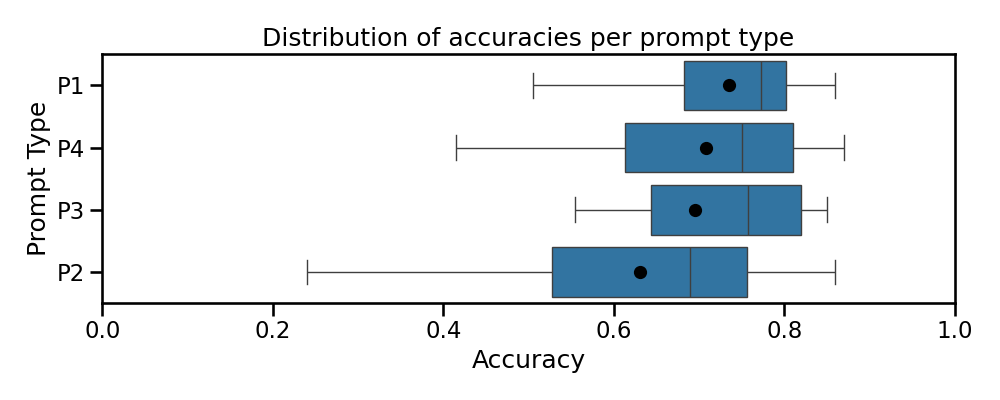}
    \caption{Accuracy distribution of the results for the UKP database per model-topic pair, broken down by prompt types, for all models}
    \label{A1}
\end{figure}

Next, the ablation studies were performed to show the impact of four defined prompts. In each experiment, one prompt was removed and the voting between the remaining three was performed. Intuitively one might suppose that removing the worst performing prompt (P2, cf. Fig. \ref{A1}), the performance of the voting algorithm would improve. However, the results as seen in table \ref{tab:ablation} point to a different conclusion: First, there is no clear 'optimal' selection of prompts. Depending on the model, different combinations of the three prompts perform best (marked in bold), with each combination claiming a 'win' at least once. Second, using all four prompts yields on average almost always a better aggregate score than choosing some subset of the prompts for the voting algorithm.

\begin{table}[H]
\centering
\caption{Ablation study of the impact of presented prompts. }
\label{tab:ablation}
\small
\begin{tabular}{lcccccc}
\toprule
model & P2,3,4 & P1,3,4 & P1,2,4 & P1,2,3 & avg & P1,2,3,4 \\
\midrule
Llama 1b      & 30,3 & 48,7 & \textbf{48,7} & 35,5 & 40,8 & 42,1 \\
Llama 3b      & 59,7 & 62,8 & 61,2 & \textbf{64,5} & 62,1 & 62,7 \\
Llama 8b      & 72,9 & 72,9 & 71,2 & \textbf{75,8} & 73,2 & 73,0 \\
Llama 3.1 70b  & 76,6 & 76,1 & 76,6 & \textbf{76,9} & 76,6 & 76,6 \\
Llama 3.3 70b  & 76,5 & 75,5 & \textbf{77,5} & 76,8 & 76,6 & 77,6 \\
DS 70b   & 80,3 & \textbf{80,7} & 79,9 & 80,5 & 80,4 & 81,7 \\
GPT4    & 81,8 & 80,8 & \textbf{81,9} & 81,7 & 81,6 & 82,0 \\
GPT4o   & \textbf{84,8} & 84,1 & 84,6 & 84,7 & 84,6 & 84,8 \\
DS R1    & 79,8 & \textbf{80,4} & 79,8 & 79,8 & 80,0 & 80,0 \\
\bottomrule
\end{tabular}
\end{table}

\begin{table}[H]
\centering
\begin{tabular}{|c|c|c|c|}
\hline
\textbf{} & \textbf{UKP (F1)} & \textbf{Args.me (Acc)} \\ \hline
GPT-4    &     79.2                                            &  88.4  \\ \hline
GPT-4o    &   \textbf{83.2}                                              &  \textbf{90.6}  \\ \hline
DS R1    &     81.0                                            &   90.4  \\ \hline
BERT (\cite{})    &       57.7                                          &    85.3 \\ \hline
LSTM (\cite{ref_stab18})    &       42.85                                          &   - \\ \hline
\cite{pietron2024efficient}    &   68.5                                              &   89.6 \\ \hline
\cite{akiki2020exploring}        &     -                                            &   75.5 \\ \hline

\end{tabular}
\caption{Accuracy comparison across different argument mining deep learning solutions.}
\label{tab:comparative}
\end{table}

Finally, in tab.\ref{tab:comparative} the comparative results are described. They show that GPT-4o with Deepseek-R1 significantly outperforms other deep learning models. It should be noted that GPT-4o achieves slightly better F1 and accuracy values than Deepseek-R1.

\begin{figure}[h]
    \centering
    \includegraphics[width=14cm]{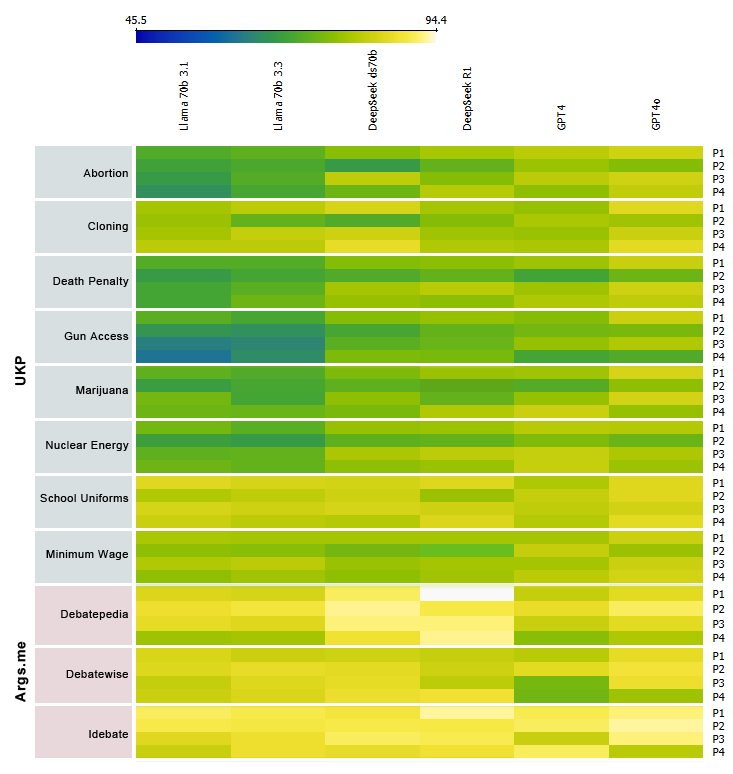}
    \caption{Heatmap visualizing the performance of the six top-performing LLM models across debate topics in the UKP and Args.me datasets.}
    \label{heatmap1}
\end{figure}

A heatmap visualizing the performance of the six best performing LLM models (Fig. \ref{heatmap1} shows that there are datasets where the models perform significantly better, approaching near-maximum accuracy. This is the case for DeepSeek-R1 on the Debatepedia dataset and GPT-4o on the Idebate dataset. However, there are also areas where the models perform notably worse—particularly the \textit{UKP gun acces}s, \textit{death penalty} and \textit{abortion} datasets, where LLaMA shows especially weak results, even though other models do not exhibit similar anomalies. Furthermore, GPT-4 performs noticeably worse in the \textit{Debatepedia} dataset.

A limited scope evaluation was also performed on the latest available GPT model, o3-mini. Due to the lack of access to the API, the evaluation was performed manually on a very small data sample: the model was tested on the UKP \textit{minimum\_wage} dataset using prompt P4 only. The choice of this particular combination of dataset and prompt was made due to the fact that the results of all the other models for this combination are closest to the means calculated for each model from all dataset-prompt combinations. 
The accuracy rate for o3-mini was 82.5\%, which means that the model was second best for this dataset-prompt combination: the best model, GPT 4o, achieved 84.5\%.

\begin{figure}[h]
    \centering
    \includegraphics[width=16cm]{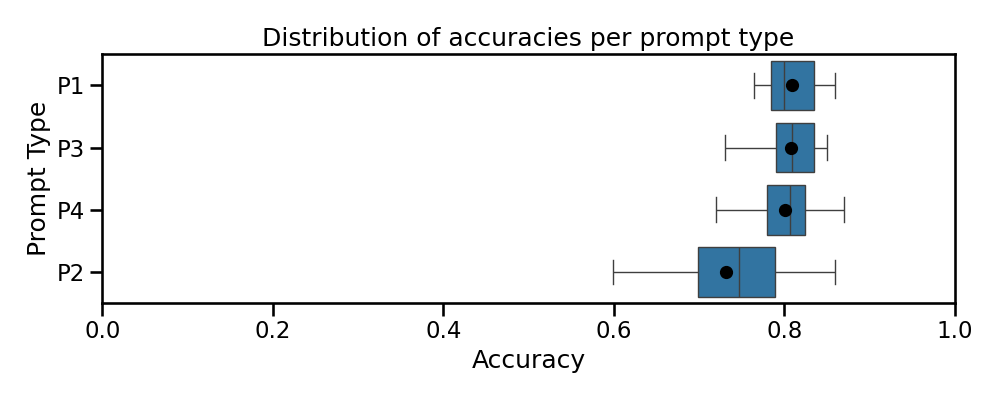}
    \caption{Accuracy distribution of the results for the UKP database per model-topic pair, broken down by prompt types, for the GPT and DeepSeek models.}
    \label{A2}
\end{figure}

Despite achieving overall good performance in argument classification, the models still make errors. Differences in the performance of individual prompts are not substantial overall. For example, when considering the average number of errors made by all models in classifying the UKP datasets (see Figure \ref{A1}), the variations appear minor. However, these differences become more pronounced when we analyze the errors made specifically by the most accurate models, such as DeepSeek and GPT-4/4o. As shown in Figure \ref{A2}), prompt P2 performs the worst by a significant margin, followed by prompt P4 as the second least effective.

As observed, prompts that require answers in natural language perform better than those where the response must be given using letter codes such as F/A/N. This may be due to the model expending additional resources to translate coded responses into words. Additionally, the weaker performance of prompt P2 suggests that, in the analyzed cases, the more simply the thesis is formulated, the better the results. Nevertheless, prompt P2 outperforms others in certain cases. For example, as shown in Figure \ref{heatmap1}, it achieves the highest accuracy in classifying data from Idebate (within the Args.me dataset). This suggests that a more elaborate formulation of the thesis can, in some cases, lead to better results.

\begin{figure}[h]
    \centering
    \includegraphics[width=16 cm]{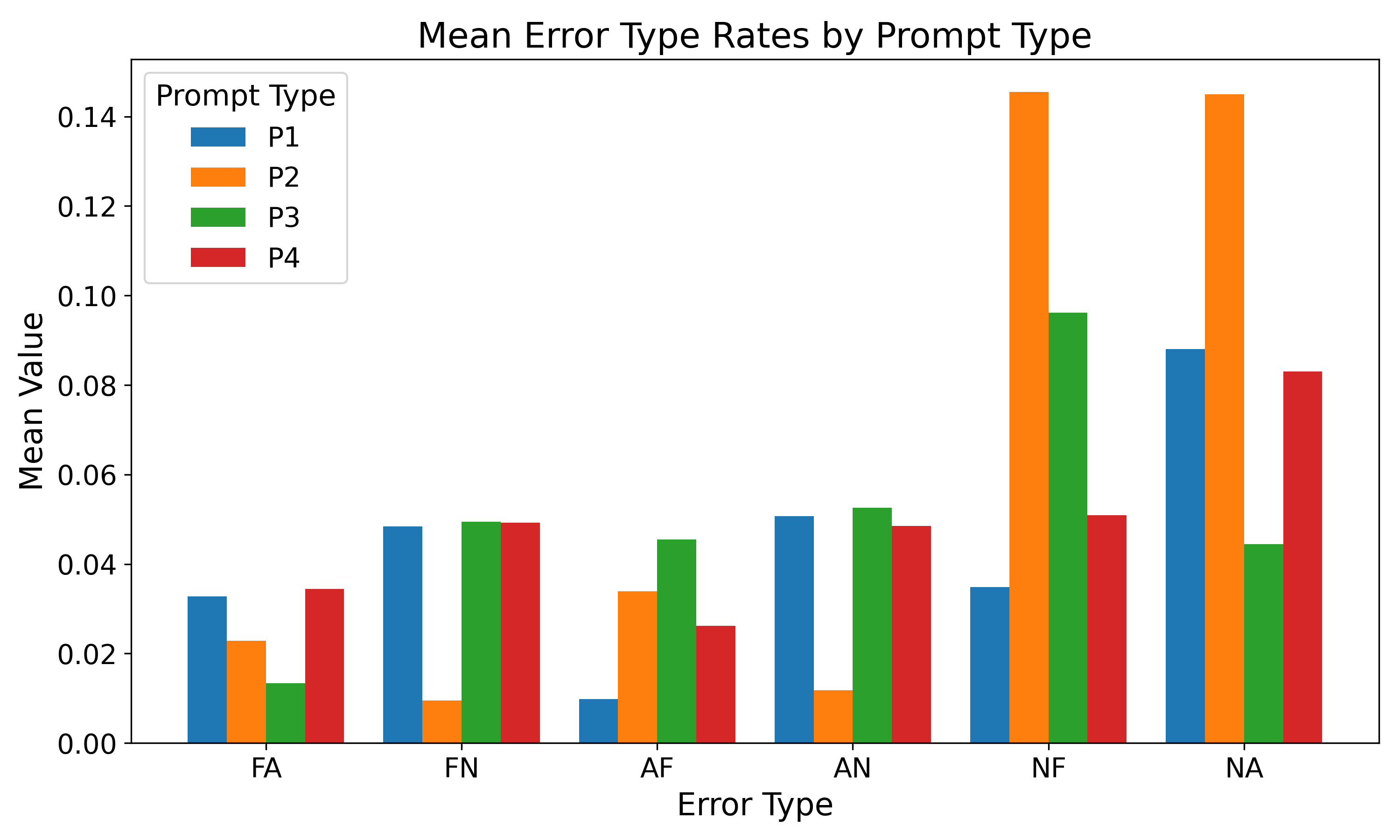}
    \caption{Percentage of each error type depending on the prompt used  (sample:UKP dataset).}
    \label{A3}
\end{figure}

Interesting patterns emerge when we analyze the types of errors that occur. Figure \ref{A3}) shows the proportion of the most common error types out of all errors within each prompt, following this classification:

\begin{itemize}
\item \textbf{AF} and \textbf{AN} refer to statements labeled by human annotators as \textit{against} but incorrectly classified by the model as \textit{for} or \textit{neutral}, respectively.
\item \textbf{FA} and \textbf{FN} refer to statements labeled as \textit{for} but incorrectly classified as \textit{against} or \textit{neutral}.
\item \textbf{NA} and\textbf{NF} represent statements annotated as \textit{neutral} but misclassified by the model as \textit{against} or \textit{for}, respectively.
\end{itemize}

The most frequent type of error made by the models is classifying neutral utterances—labeled as such by annotators— making them arguments. This suggests that the prompt’s directive to find arguments may lead the models to overinterpret content in search of argumentative structure. All prompts tend to produce NA errors—that is, misclassifying neutral statements as counter-arguments. Prompt P2, in particular, exhibits a significantly stronger tendency than the others to mislabel neutral utterances as supporting arguments (NF errors). In contrast, prompt P3 shows the most balanced distribution of error types, indicating no strong bias toward any specific classification error.

\begin{figure}[h]
    \centering
    \includegraphics[width=16 cm]{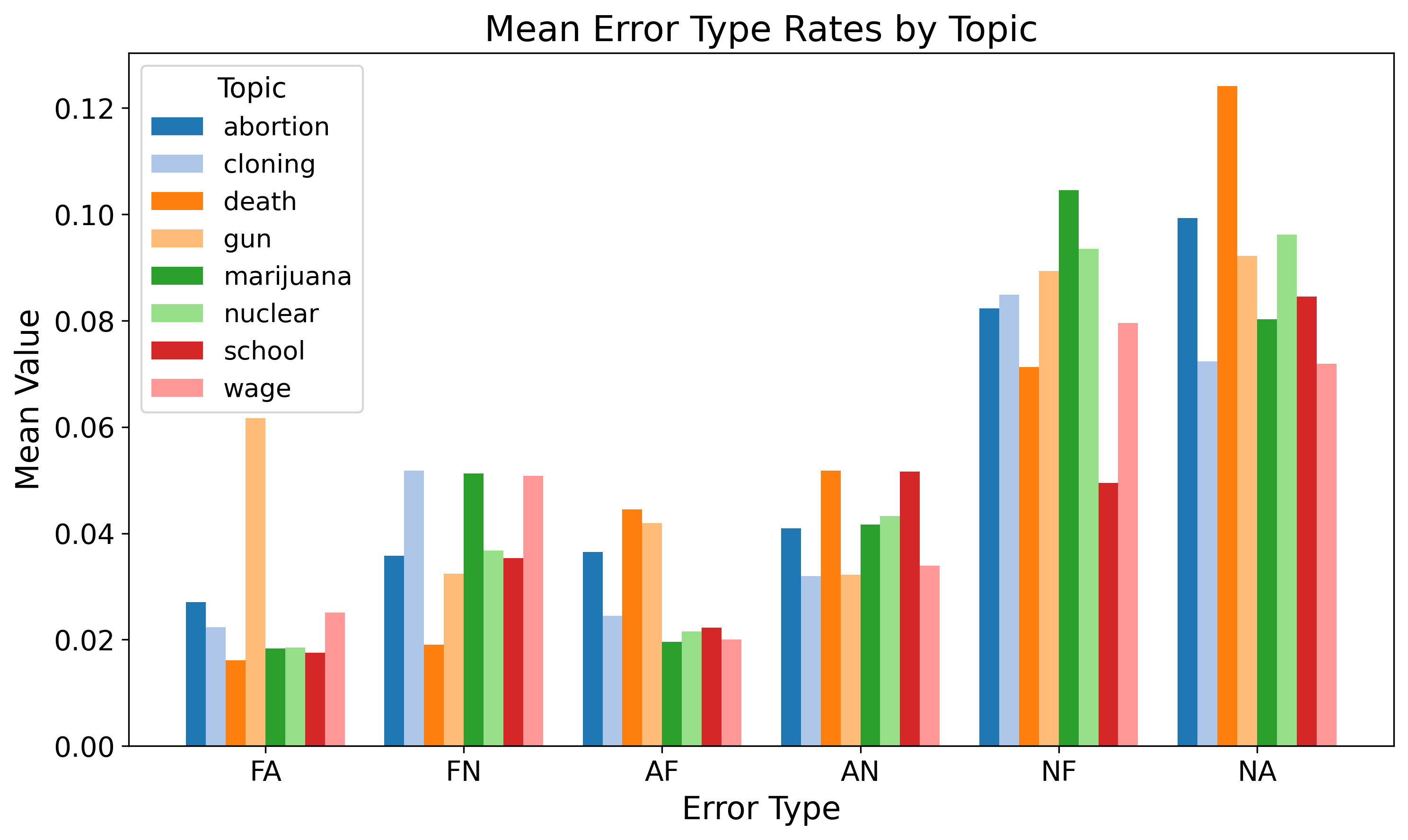}
    \caption{Percentage of each error type by debate topic in the UKP dataset.}
    \label{A4}
\end{figure}

Figure \ref{A4} presents an analysis of error types by debate topic in the UKP dataset, which reveals several noteworthy patterns. When it comes to the misclassification of posts labeled as neutral by human annotators, the topics most prone to being incorrectly classified as counter-arguments (NA errors) are \textit{death penalty} and \textit{school uniforms}. In contrast, the topics most susceptible to NF errors — where neutral statements are misclassified as supporting arguments — are \textit{marijuana} and \textit{cloning}. The topic of \textit{gun access} stands out due to a notably high frequency of both FA and NA errors — where pro-access and neutral statements are misclassified as opposing arguments. This makes it the topic most prone to false negatives and suggests potential bias in the models' training data.

\begin{figure}[h]
    \centering
    \includegraphics[width=16 cm]{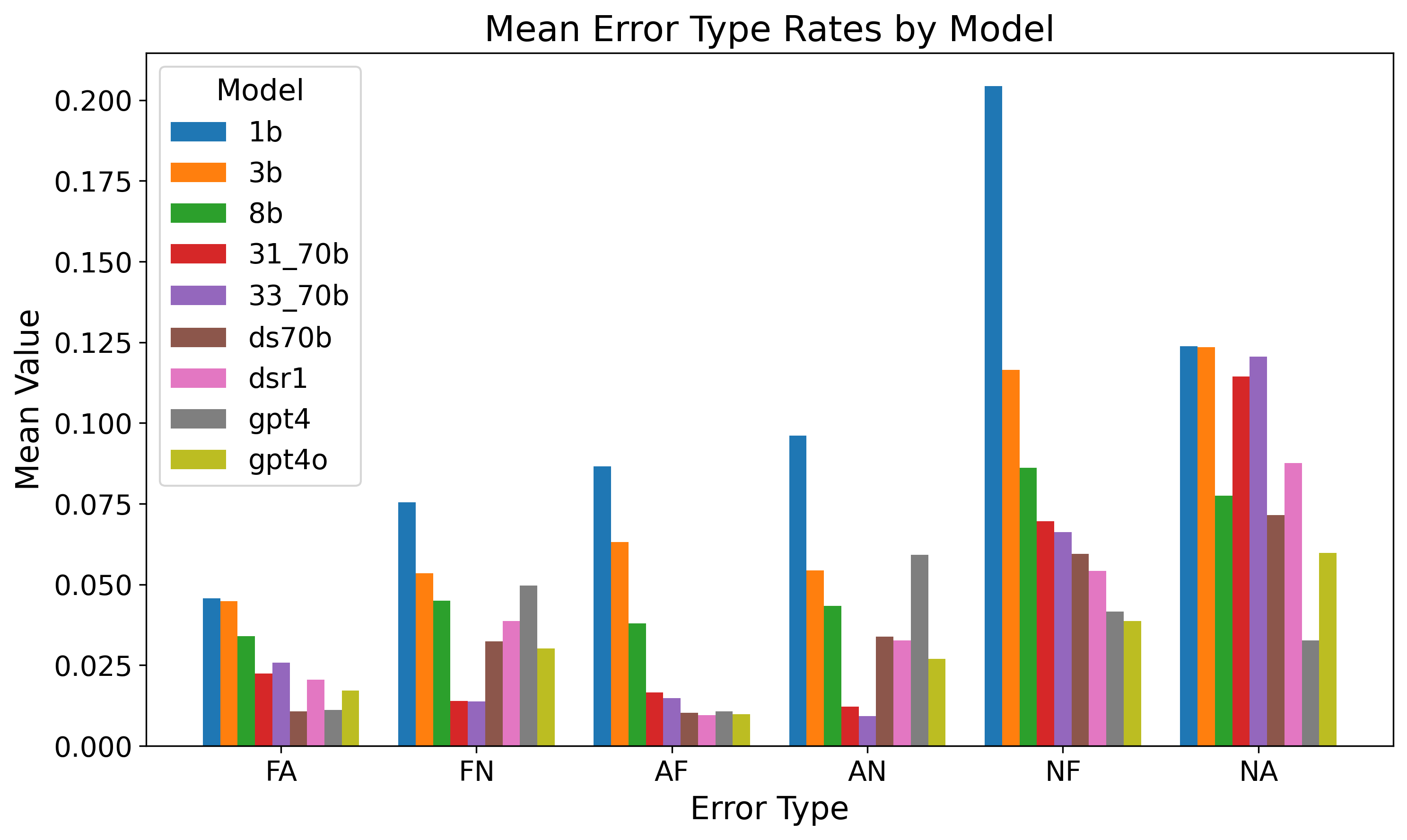}
    \caption{Percentage of each error type by debate topic in the UKP dataset.}
    \label{A5}
\end{figure}

In turn, when we examine the distribution of error types by model, as visualized in Fig. \ref{A5}, we see that GPT-4 stands out from the other models, showing a tendency to misclassify arguments as neutral statements (FN and AN errors).

In order to examine the nature of errors made by the models and the contexts in which they occur, we conducted a qualitative analysis of selected error types. As the subject of detailed analysis, we focused in particular on the errors made by the GPT-4 and LLaMA 70B models, as they achieve relatively strong results but are still prone to certain misclassifications. The findings of this analysis are presented below:

\begin{enumerate}
    \item \textbf{Literal adherence to the facts presented in statements without recognizing the intention behind them.} For example, regarding the claim "Nuclear Energy should be developed," the sentence "the decommissioning process costs between US \$300 million to US \$5.6 billion" was classified by human annotators as an argument against, because it highlights the significant financial burden associated with the decommissioning of nuclear power plants. This information underscores one of the major economic challenges of nuclear energy: the high costs not only of building and maintaining nuclear power plants but also of safely decommissioning them at the end of their lifecycle. However, GPT 4 and Llama in some cases classified this sentence as a "non-argument," reasoning that it just provides factual information about the costs associated with decommissioning nuclear power plants without directly supporting or opposing the development of nuclear energy. Similarly, the emotionally charged statement against the death penalty: "I went in an innocent man and I came out an innocent man" was considered by GPT 4 as not an argument, stating that it is rather a reflection of personal experience than a direct argument regarding the policy or ethical considerations of the death penalty.
\item \textbf{High use of negations within an argument can sometimes lead GPT 4 and Llama to misunderstand the meaning of a statement.} For example, the sentence "The fact that some states or countries which do not use the death penalty have lower murder rates than jurisdictions which do is not evidence of the failure of deterrence" was correctly classified by human annotators as an argument supporting the death penalty. However, GPT 4 misinterpreted it as an argument against the death penalty, arguing that the correlation between lower murder rates and the absence of the death penalty does not necessarily prove that the death penalty serves as an effective deterrent to crime. In other words, GPT 4 overlooked in its interpretation the final negation present in the argument, concluding that this statement is evidence of the failure of deterrence.
\item \textbf{Incorporating context that was not present in the analyzed content but is often found in public debate.} For instance, regarding the thesis "cloning should be allowed," the statement "DNA cloning has been used in genetic engineering to create plants that offer better nutritional value" referenced a specific application of DNA cloning in the context of agriculture, providing an argument in favor of allowing cloning (at least in this area). However, GPT 4 concluded that this statement does not explicitly address the ethical, legal, or social implications of cloning in the context of cloning humans or animals, and therefore, categorized it as not an argument. This occurred even though GPT 4 was not asked to consider the context of humans or animals, but rather to address cloning in general. It seems that the model introduced into the general issue of cloning the often-discussed ethical, legal, or social implications of cloning humans or animals.
\item \textbf{Failure to understand simple, emotional arguments expressed in everyday language.} For example, Llama determined that the sentence "Cloning can help save lives" does not present a clear argument for or against the permissibility of cloning, as it does not provide any reasoning or evidence to support a specific position on the proposition. However, human annotators clearly recognized this as an argument in favor of cloning.
\item \textbf{Inconsistencies in interpreting complex and multi-faceted arguments.} For instance, in relation to the thesis "Gun access should be limited," an argument was made against excessive restrictions on firearm access imposed on individuals who commit minor economic offenses. When elaborating on its interpretation, GPT 4 initially claimed that this is an argument against limiting access to firearms, only to later conclude that it is, in fact, an argument against firearm access.
\item \textbf{Loss of reference to the specific thesis being addressed,} resulting in misclassification of whether a statement supports or opposes the given claim. This often occurs when the thesis is phrased in a more complex or nuanced way. For example, in one case, the thesis is: \textit{There is a correlation between the laxity of a country’s gun laws and its suicide rate.} The accompanying premise challenged this assumption by arguing that while there may indeed be a correlation with gun access, it relates not to suicide rates but rather to increased safety and civic responsibility. The GPT 4o model correctly understood the contradiction between the thesis and the argument but misclassified the stance, labeling the premise as supportive, likely because it favored gun access in general, even though it contradicted the specific claim about suicide.
\end{enumerate}

We observed earlier that certain topics were significantly more prone to classification errors. In the case of the death penalty debate, some models particularly struggled with arguments supporting its application. An extreme example of this phenomenon occurs with LLaMA 3.3 70B using prompt P4—within the group of pro–death penalty arguments, as many as 53.8\% were misclassified (as FA or FN), while in the opposing group, errors were far less frequent (only 5.6\% misclassifications of AF and AN type). Moreover, when misclassifications involved statements labeled as "no argument," nearly all of them were interpreted by LLaMA as opposing the death penalty. This suggests the presence of potential bias in the model when processing content related to this topic.

Another interesting case was the topics of abortion. In this instance, Llama generally made more classification errors than in other topics and did so more often than GPT-4. A common error in this debate was attributing meaning beyond the original text. For example, the general statement \textit{Every parent wants the best for their child} was interpreted by Llama as an argument against abortion, even though neither the annotators nor GPT-4 decided so. Conversely, the fairly straightforward sentence "Some abortions happen because of societal pressures," which GPT-4 correctly identified as an argument against abortion, pointing out that some women are pressured into abortions against their will, was for unknown reasons treated by Llama as an argument in favor of abortion.


\subsection{Comparison of GPT-4o and DeepSeek R1 using examples of misclassification}

When analyzing the two models that perform the best, GPT-4o and DeepSeek R1, we observe that although DeepSeek R1 often achieves results similar to GPT-4o, its performance drops significantly in certain configurations. As shown in Fig. \ref{heatmap1}), there are specific topics and prompts where DeepSeek R1’s accuracy falls behind other models, indicating that its behavior is inconsistent.

A detailed examination of the classification errors made by both models reveals that GPT-4o is better at interpreting pragmatic relationships and the logical structure of arguments. It is also more robust against misleading linguistic features such as negation, irony, and counterfactual reasoning, which occasionally confuse DeepSeek R1.
One illustrative case involves arguments that appear to support a given claim but include warnings or risks, which can mislead DSR1. In the Debatepedia (Args.me) dataset, a claim states: “There is a risk that iron fertilization could result in increased production of nitrous oxide, a greenhouse gas far more powerful than carbon dioxide.” One of the related arguments is: “Iron-fertilized algae blooms may release some greenhouse gases (e.g. nitrous oxide, a potent greenhouse gas).” Human annotators correctly classified this as supporting the claim. However, DSR1 labeled it as opposing, likely misled by the conditional language (“may release”) and the mention of a negative effect in what is essentially a pro argument.

Another example comes from the Debatewise (Args.me) dataset, which includes the claim: “If we have been to the moon, surely we could go back”—a reference to conspiracy theories questioning the authenticity of the U.S. moon landing. One supporting argument consists of a series of rhetorical questions (e.g., “Why did Russia not follow?”; “Why has the USA stopped going to the moon if it can?”). DeepSeek R1 misinterprets the structure of this argument and incorrectly classifies it as opposing.

Despite its generally high accuracy, DeepSeek R1’s performance deteriorates on certain topics and prompt types. In contrast, GPT-4o handles pragmatic relations, complex logical structures, and subtle linguistic cues—such as negation or counterfactual reasoning—with near-perfect consistency. Examples from the Debatepedia and Debatewise datasets demonstrate that DSR1 is prone to misclassifying conditionals and rhetorical constructions that require deeper interpretation of speaker intent.

On the other hand, GPT-4o occasionally struggles with detecting implicit criticism. For example, it failed to recognize that the argument: "Gun companies fund elections. Leaders in America are backed by people who profit from gun sales" actually undermines the thesis that a “new era” in U.S. governance is something positive. Similarly, in response to the thesis: “Iraq occupation of Kuwait would have driven up oil prices, hurting the U.S. economy,” GPT-4o misclassified a lengthy supporting argument—demonstrating the destructive impact of the invasion on both the Middle East and the U.S. economy—as opposing the thesis. This likely resulted from the model losing track of the argument’s context and speaker’s intent.

Despite such errors, the models tend to complement one another:
DSR1 excels at literal interpretation, whereas GPT-4o performs better with complex linguistic constructions. Summarizing the differences:

\begin{itemize}
    \item DSR1 makes errors in a more systematic manner, primarily due to its limitations in natural language understanding. These mistakes typically stem from a lack of comprehension of complex sentence structures, contrastive phrasing, and modal expressions.

    \item GPT-4o’s errors are less systematic and more context-dependent. When it does make mistakes, they often relate to pragmatics—such as failing to detect hidden criticism or implicit intent. It processes language more effectively overall but may struggle with nuanced interpretation.
\end{itemize}

The examples of error types made by both models are summarized in Table \ref{tab:errors} below.

\begin{table}[H]
\centering
\caption{Examples of error types made by DeepSeek R1 and GPT-4o.}
\label{tab:errors}

\renewcommand{\arraystretch}{1.4} 

\begin{tabular}{|p{3cm}|p{6cm}|p{6cm}|}
\toprule
\textbf{Aspect} & \textbf{DSR1} & \textbf{GPT-4o} \\
\midrule
\textbf{Error Style} & Systematic, recurring patterns & Varied, often incidental, linked to interpretation errors \\
\textbf{Sensitivity to Syntax} & Sometimes struggles with contrasts (e.g., “but”, “however”), misses punchlines or key clauses & Better at detecting semantic shifts and contrasting clauses \\
\textbf{Factual Logic} & Usually correct but tends toward simplified analysis	 & Occasionally fails in causal reasoning or contextual interpretation \\
\textbf{Implicit Criticism} & Typically able to recognize & Sometimes fails to detect the speaker’s critical stance \\
\bottomrule
\end{tabular}

\renewcommand{\arraystretch}{1} 
\end{table}

\subsection{Errors made by LLMs vs. errors made by annotators}

The accuracy of the models was assessed based on the discrepancies between their evaluations and those made by the annotators who created the datasets used in this research. In other words, the values presented in the tables actually represent the degree of agreement between the models’ and the annotators’ judgments. 
But is it not possible that, in some cases, these discrepancies are due not to errors made by the models but rather by the annotators themselves? To fully answer this question, one would need to carefully examine all the records from the datasets, which would require a significant amount of work. 
Instead, a preliminary investigation was conducted, focusing on records from the UKP \textit{minimum wage} dataset where, using prompt P4, the best-performing models—GPT-4o and o3-mini—produced responses that differed from those of the annotators. For the o3-mini model, there were 35 such records, and for GPT-4o, there were 31. 
It is worth noting that these sets overlap on 20 records, meaning that in a considerable number of cases, the models disagree with each other: o3-mini has 15 unique discrepancies with the annotations, while GPT-4o has 11.

After reviewing all 46 records, it turned out that a significant portion of the models’ so-called “incorrect” responses were actually due to annotator errors. For the o3-mini model, there were 10 such cases, accounting for 28.6\% of its “incorrect” responses, while for the GPT-4o model, there were 13 cases, or as much as 41.9\%. Additionally, in a few instances, both the annotators’ and the models’ responses were incorrect—that is, the correct answer was a third option. There were 4 such cases for o3-mini (11.4\% of all its “incorrect” responses) and 1 case for GPT-4o (3.2\%).

In both cases studied, most of the records that disagreed with the annotations were still due to model errors, although it should be noted that the GPT-4o model performed significantly better in this regard. This was largely because it was more “cautious”: it more frequently classified records as “no argument,” which often turned out to be correct. This was usually because the content of the expression—although it had an argumentative character—when taken out of the context of the discussion, did not clearly indicate whether it was used as an argument for or against the thesis.

In light of this data, one might conclude that the actual accuracy of the models is higher—91\% for GPT-4o and 87.5\% for o3-mini. However, this would be a premature conclusion, as it is not known whether the annotators also made mistakes in cases where the models agreed with them—such cases would have to be counted as actual model errors. One can only speculate that such situations are very rare or possibly nonexistent, since agreement cases are likely easier to classify than the disagreement cases that were analyzed. However, such speculation does not provide a reliable basis for any definitive numerical claims.



\section{Conclusions and future work}

Our analysis demonstrates that while GPT-4/GPT-4o, Llama and Deepseek-R1 represent a significant step forward, there is still room for improvement, especially in dealing with more subtle aspects of language and argumentation. These observations contribute to ongoing discussions about the reliability of AI in more sophisticated, real-world applications, and they are crucial for future model enhancements.
Errors in logical reasoning and improper responses to certain rhetorical expressions are well-documented and persistent challenges for LLMs (\textcite{dentella2024testing}; \textcite{murphy2025fundamental}). Many indications suggest that, in transformer-based technology, these issues may never be fully eliminated.

The presented research provides a basis for building better data sets containing fully correct arguments and annotations about the strength of a given argument. These observations suggest that greater effort should be made to create higher-quality datasets to reduce or eliminate annotation errors. Additionally, it would be beneficial to publish the annotation policy adopted during dataset creation, specifying which borderline categories of statements are considered arguments and which are not.



Based on the results and observations, future work might focus on developing more sophisticated prompt engineering-based algorithms that improve argument classification in the reasoning process. A plausible next step would be to adapt the RAG technique to support the argument classification process.

\printbibliography
\appendix
\section{Example sentences for few shot prompt}
\begin{itemize}
\item \textbf{Abortion}:
    
    \textbf{for}: 'The choice — the only actual choice , in the world as it really is — is between safe , legal abortion and dangerous , illegal abortion .'
    
    \textbf{against}: 'In this case we may never do evil ( directly attack and kill a child via abortion ) so that good ( saving the life of the mother ) may result .'

    \textbf{no argument}: 'This means it has to steer monetary policy to ( a ) keep prices stable , and to ( b ) keep unemployment low and the economy growing .'

\item \textbf{Cloning}:
    
    \textbf{for}: 'These alternatives not only avoid the ethical problems inherent in using human embryos but have also been more successful to date .'
    
    \textbf{against}: 'Sentence: Senator Landrieu , a supporter of abortion choice , argues that cloning is too unreliable  .'

    \textbf{no argument}: 'So ... is cloning good or bad ?'

\item \textbf{Death penalty}:
    
    \textbf{for}: 'In 1976 , the Supreme Court moved away from abolition , holding that " the punishment of death does not invariably violate the Constitution .'
    
    \textbf{against}: 'Opponents of retribution theory believe in the sanctity of life and often argue that it is just as wrong for society to kill as it is for an individual to kill .'

    \textbf{no argument}: 'ProCon.org has more than 20,000,000 annual readers .'

\item \textbf{Marijuana}:
    
    \textbf{for}: 'Eight people or groups turned in arguments for the " pro " side 's six pages in the publication , urging voters to approve the Regulation and Taxation of Marijuana Act ( RTMA ) .'
    
    \textbf{against}: 'She likens it to making alcohol or cigarettes ten times stronger .'

    \textbf{no argument}: 'The state is enjoying economic growth and the lowest unemployment rate in years .'

\item \textbf{Gun control}:
    
    \textbf{for}: 'According to John R. Lott Jr. , PhD , " when states passed concealed carry laws during the 19 years we studied ( 1977 to 1995 ) , the number of multiple-victim public shootings declined by 84\%'
    
    \textbf{against}: 'Education Is The Answer More harsh gun control laws are not needed .'

    \textbf{no argument}: '“ I had deep anger when I heard that , ” he told me .'

\item \textbf{Minimum wage}:
    
    \textbf{for}: 'Milton Friedman called them a form of discrimination against low-skilled workers'
    
    \textbf{against}: 'Not true : The typical minimum wage worker is not a high school student earning weekend pocket money .'

    \textbf{no argument}: 'If the crabs were to work together , they could all easily escape .'

\item \textbf{Nuclear energy}:
    
    \textbf{for}: 'Fossil fuels receive large direct and indirect subsidies , such as tax benefits and not having to pay for the greenhouse gases they emit .'
    
    \textbf{against}: 'The number of U.S. reactors shut down for a year or longer to address numerous safety problems provides strong evidence of poor safety practices and inadequate NRC enforcement .'

    \textbf{no argument}: 'And long-term programmes to build alternatives help lay the basis for future anti-nuclear campaigns .'

\item \textbf{School uniforms}:
    
    \textbf{for}: 'People who are for uniforms say that it promotes social conformity , so the less-to-do do n’t have to be pressured to keep up with their well-off contemporaries .'
    
    \textbf{against}: 'Because children are constantly growing , there is a captive market for new school clothes and manufacturers take advantage .'

    \textbf{no argument}: 'But I think this , I think that local communities and states should make the decision and I feel very strongly about that .'

\end{itemize}

\end{document}